\documentclass[11pt]{article}

\usepackage[T1]{fontenc}
\usepackage{mathptmx}
\usepackage{amsmath}
\usepackage{amssymb}
\usepackage{amsfonts}
\usepackage{graphicx}
\usepackage{booktabs}
\usepackage[round]{natbib}
\usepackage{authblk}
\usepackage{setspace}
\usepackage{fancyhdr}
\usepackage{hyperref}

\title{Using Human-LLM Disagreement to Improve Checklist-Based Quality Appraisal}

\author[1]{Timo van der Kuil}
\author[1]{Bruno Messina Coimbra}
\author[2]{Mirjam van Zuiden}
\author[1]{Robert A. Bagheri}
\author[1]{Rens van de Schoot}
\author[1]{Klaas Dieleman}
\author[1]{Berend Greijn}
\author[1]{Stefan Houkes}
\author[1]{Sebastiaan Rodenhuis}
\author[1]{Elizabeth M. Grandfield}

\affil[1]{Methodology and Statistics, Utrecht University, Utrecht, the Netherlands}
\affil[2]{Clinical Psychology, Utrecht University, Utrecht, the Netherlands}

\date{}
\begin{document}
\maketitle

\begin{abstract}Systematic reviews rely on quality appraisal of included studies, a process that is time-consuming and sensitive to ambiguity in checklist criteria. Although large language models (LLMs) offer opportunities to support these tasks, appraisal checklists are typically treated as fixed inputs, and it remains unclear how their design affects agreement with expert judgments. Therefore, we investigate (1) whether LLMs can approximate human judgments in checklist-based appraisal and (2) whether patterns of human–LLM disagreement can be used to identify and improve ambiguous checklist items. Using the Guidelines for Reporting on Latent Trajectory Studies (GRoLTS) checklist, we compare LLM-generated assessments with expert annotations across three research topics and two checklist versions. Agreement is assessed using item-level accuracy, chance-corrected agreement, and preservation of study-level rank ordering.

We find that performance varies substantially across checklist items, with ambiguous and conditional criteria producing the greatest disagreement. Revising these items improves both raw and chance-corrected agreement. Although item-level misclassifications persist, LLM-generated scores often preserve the relative ranking of studies when high-agreement items are retained. These results indicate that reliable LLM-assisted appraisal depends not only on model choice but also on checklist design. The findings suggest that analyzing human–LLM disagreement can help identify problematic checklist items and support the iterative improvement of research synthesis workflows.

All code and model outputs are publicly available at \url{https://github.com/timovdk/grolts-llm}.
\end{abstract}

\section{Introduction}

The volume of published research continues to grow rapidly, making research synthesis increasingly complex and time-consuming~\citep{thelwall2022, bornmann2015}. Beyond identifying relevant studies, systematic reviews require researchers to assess the quality of each included study~\citep{higgins2024}. This process, even when guided by structured checklists or standardized forms, remains labor intensive and cognitively demanding. Estimates based on questionnaire responses suggest that quality assessment alone can take 30 to 60 minutes per study~\citep{haddaway2019} and may be susceptible to fatigue related errors, particularly in large reviews~\citep{Wang2020-uy}. As a result, quality assessment is a substantial portion of the manual effort involved in systematic reviews, highlighting the potential value of (semi)automated tools that can support this task. At the same time, the reliability of quality appraisal does not only depend on the effort of the reviewer but also on the clarity and structure of the appraisal instruments themselves.

Recent advances in large language models (LLMs) have created new opportunities to support these labor intensive stages of the systematic review process. Earlier automated systems such as RobotReviewer~\citep{marshall2017} demonstrated that methodological characteristics could be automatically extracted from clinical trial reports, providing machine-generated risk-of-bias assessments that support human reviewers. Building on these earlier efforts, a growing body of work has explored the use of LLMs in evidence synthesis workflows, including title-abstract and full-text screening~\citep{vandeschoot2025, Dennstadt2024, Guo2024, bron2024}, data extraction~\citep{Li2026, Khraisha2024}, and risk-of-bias assessment~\citep{Hasan2024}. Several studies report promising performance. For example,~\citet{Dennstadt2024} and~\citet{Guo2024} report screening accuracies approaching 90\%, although performance varies between datasets and review topics. In data extraction tasks,~\citet{Li2026} show that LLMs can achieve high precision (between 0.75 and 0.95), but often suffer from lower recall because relevant information is omitted. Similarly,~\citet{Khraisha2024} find that LLM performance can approach that of human reviewers for certain tasks under specific conditions. These findings suggest that LLMs can support several stages of the review process, although performance varies between domains, LLMs, and prompt designs.

Methodological and reporting quality appraisal presents particular challenges for automation. Checklist-based assessments require reviewers to interpret methodological descriptions and apply criteria that may involve implicit assumptions or domain-specific judgment. This challenge is illustrated by~\citet{Hasan2024}, who report overall risk-of-bias agreement of 61\% between humans and LLMs. Empirical studies of appraisal tools show that even human reviewers often disagree when applying the same criteria, with inter-rater variability frequently attributed to ambiguous wording or underspecified checklist items~\citep{Hartling2013, ArmijoOlivo2012, Higgins2011}. For example, reviewers may differ in how they interpret terms such as ``adequate'' follow-up or ``appropriate'' statistical adjustment. Double-barreled items such as ``Were confounders identified and controlled for?'' may further contribute to variability if assessors focus on only one component of the criterion. Similar challenges arise when LLMs are used for appraisal tasks. Because they rely on probabilistic next-token prediction rather than fixed rules, subtle variations in phrasing can influence the interpretation and thus the resulting judgment~\citep{Chang2024, anagnostidis2024}. Furthermore, LLMs can generate hallucinations, posing risks in settings where factual precision and transparency are essential~\citep{Huang_2025, bommasani2022}. 

Despite growing interest in LLM-assisted evidence synthesis, appraisal instruments are typically used as fixed inputs. For example,~\citet{Hasan2024} apply LLMs to the ROBINS-I risk-of-bias tool without modifying its structure or wording, instead exploring how different prompts and models affect performance. Across studies, experimental variation tends to focus on these aspects, while the checklist items themselves remain unchanged~\citep{Khraisha2024, Li2026}. As a result, it remains unclear whether modifying checklist items, for example by clarifying ambiguous phrasing, separating conditional criteria, or simplifying complex questions, could improve agreement between human and LLM judgments. Furthermore, it is not known whether LLM-generated checklist scores preserve the relative order of studies according to reporting quality.

In this study, we compare LLM and human judgments at the item level to identify sources of disagreement by evaluating LLM-assisted checklist-based appraisal using the Guidelines for Reporting on Latent Trajectory Studies (GRoLTS)~\citep{vandeSchoot2017}, a widely used and expert-designed reporting checklist aimed at improving transparency and consistency in studies employing latent trajectory models. We implement a retrieval-augmented generation (RAG) annotation pipeline that enables LLMs to apply checklist items directly to full-text research articles. Retrieval-augmented approaches have shown strong performance in domain-specific question-answering systems, including applications in legal reasoning and radiology reporting~\citep{wiratunga2024, tayebi2025}. Using a systematic review dataset of post-traumatic stress disorder (PTSD) studies annotated by human experts~\citep{vandeSchoot2017}, we compare LLM and human judgments at the item level to identify sources of disagreement, such as misinterpretation of conditional criteria or difficulty resolving incomplete contextual information. Based on these analyses, we develop a revised version of the checklist designed to reduce ambiguity, simplify complex or conditional items, and improve interpretability for both human and automated annotators. The revised checklist is subsequently applied to the PTSD dataset (remapped to the revised items where possible) and to two additional domains: adolescent delinquency and educational achievement, to investigate generalizability.

Our evaluation focuses on three complementary questions. First, we assess the item-level agreement between human experts and multiple LLMs (GPT-5-mini-2025-08-07~\citep{openai_gpt}, LLaMA 3.3 70B Instruct~\citep{meta_llama3_3_70b_instruct}, Qwen3 30B A3B Instruct 2507~\citep{qwen3technicalreport}, Qwen3-Next 80B A3B Instruct~\citep{qwen3technicalreport}, and Magistral Small 2509~\citep{mistralai2025magistral}) using accuracy, examining how performance varies by question and checklist version. Second, we use chance-corrected agreement metrics, such as Cohen's $\kappa$~\citep{cohen1960} and Fleiss' $\kappa$~\citep{Fleiss1971}, to determine whether observed agreement reflects meaningful alignment rather than random agreement or class imbalance. Third, we test whether LLM-generated checklist scores preserve the relative order of studies according to reporting quality, which is critical when appraisal scores are used to prioritize or weight studies in evidence synthesis. Together, these analyses enable us to (1) identify checklist items that are reliably automatable versus those that remain dependent on human judgment, (2) evaluate how checklist design influences human-LLM alignment, and (3) assess rank preservation as a practical criterion for LLM-assisted evidence synthesis.

The remainder of this paper is structured as follows. Section 2 describes the GRoLTS checklist, the annotated datasets, the RAG pipeline for LLM-based appraisal, and the evaluation framework, including item-level accuracy, chance-corrected agreement metrics, and rank-based analyses. Section 3 presents the empirical results, including checklist revision and cross-domain validation. Section 4 discusses implications for checklist design and LLM-assisted research synthesis, as well as limitations and directions for future research.

\section{Methods}

\subsection{Data and Materials}

\subsubsection*{GRoLTS Checklist}
The GRoLTS checklist was introduced to enhance transparency and consistency in reporting latent trajectory studies \citep{vandeSchoot2017}. It consists of 21 binary items (see Appendix~\ref{appendixA}), each scored \texttt{yes} or \texttt{no}. The total number of \texttt{yes} responses serves as a summary measure of reporting quality. The checklist was originally developed for human experts and not for automated use.

\subsubsection*{Case Study Datasets}
Three annotated datasets were used, each with a different topic: PTSD, educational achievement, and adolescent delinquency. All datasets consist of studies applying latent growth mixture modeling (LGMM) or latent class growth analysis (LCGA), which are two common latent trajectory modeling methods, used to identify unobserved subgroups in longitudinal data~\citep{Muthen2000}. LGMM allows for within-class variability, whereas LCGA assumes homogeneity within classes. These methods are used in domains such as psychology, education, and public health to discover heterogeneous patterns in trajectories.

\paragraph{PTSD}
The PTSD dataset was taken from the original GRoLTS study \citep{vandeSchoot2017} and is publicly available through the Open Science Framework (\url{https://osf.io/vw3t7/}). The dataset consists of 38 previously annotated studies. The original human expert ratings were reused without modification.

\paragraph{Educational Achievement}
The Educational Achievement dataset (\url{https://osf.io/kw8a7}) was created for the current study by retrieving studies from OpenAlex using a query requiring inclusion of ``Latent Growth Mixture Model'' or ``Latent Class Growth Analysis'' and the term ``achievement''. This search yielded 371 records. After screening based on predefined inclusion criteria (use of LGMM/LCGA, trajectories of educational achievement, and identification of latent classes), 32 studies were retained in the final corpus. All included studies were annotated using the GRoLTS checklist by a domain expert. Before annotation, the expert calibrated the interpretation of the checklist by annotating five studies from the PTSD corpus and comparing them with the original annotations. The problems encountered during annotation were discussed with a second expert and resolved by consensus. Annotation took approximately 20 minutes per study.

\paragraph{Adolescent Delinquency}
The Adolescent Delinquency dataset (\url{https://osf.io/kw8a7}) was created for the current study by retrieving studies from OpenAlex using a query combining terms related to delinquent or externalizing behavior with LGMM/LCGA terminology, restricted to publications from 2015 onward. The search yielded 407 records. After screening based on predefined inclusion criteria (use of LGMM/LCGA, adolescent externalizing trajectories, and identification of latent classes), 65 studies were retained. Of these, 37 were used in the experiments. All included studies were annotated using the GRoLTS checklist following the same calibration and consensus procedure as described for the educational achievement corpus.

\subsection{Checklist Revision Procedure}
Following the initial evaluation of human–LLM agreement in the PTSD domain using the original GRoLTS checklist (v1; see Results), the checklist was revised to reduce ambiguity and increase usability for both human and LLM annotation. The revised version (v2; see Appendix~\ref{appendixB}) was developed based on preliminary LLM outputs, systematic analysis of item-level disagreement between human and LLM annotations, and expert feedback. Disagreement patterns were examined to identify recurring sources of ambiguity (e.g., conditional phrasing, underspecified criteria, and partial reporting), which were then addressed through targeted revisions. The revisions followed four general principles:

\begin{enumerate}
    \item \textbf{Reduction of ambiguity: }Replace vague or underspecified terms with concrete, observable criteria.
    \item \textbf{Removal of double-barreled questions: }Separate items that implicitly ask multiple questions.
    \item \textbf{Removal of conditional sub-items: }Avoid ``if–then'' follow-up questions that require inferential reasoning.
    \item \textbf{Alignment with explicitly reported information: }Restrict items to content reasonably expected in the manuscript text.
\end{enumerate}

The proposed revisions were discussed in a structured feedback session at the FORAS Summit~\citep{Coimbra2025} on June 19, 2025 with eight experts who provided input on ambiguity and practical usability. Following this consultation, the revised checklist (v2) was finalized and applied in the second stage of the study across all domains. Each revised item in v2 was mapped to its corresponding item in v1 to enable comparison between human annotations collected using v1 and LLM-generated responses using v2.

\subsection{LLM-Based Annotation Pipeline}
\begin{figure}
    \includegraphics[width=\linewidth, alt={Flowchart with five sequential stages connected by arrows: Document Acquisition and Preprocessing, Text Segmentation, Embedding (showing numbered text chunks converted to vector representations), Retrieval (showing a query vector compared against stored vectors to select the top-k most similar chunks), and Prompt Construction and Answer Generation.}]{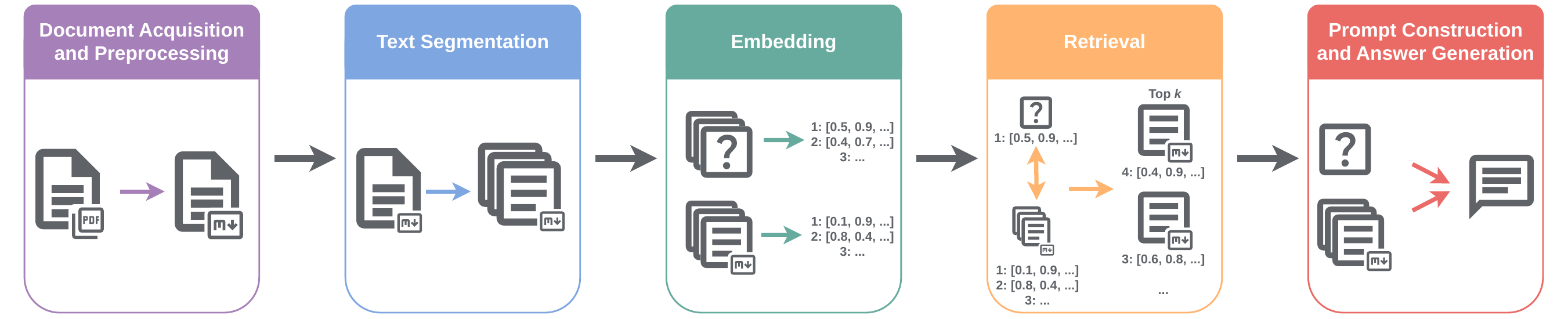}
    \caption{Overview of the five stages in the LLM-based checklist annotation pipeline. Arrows indicate the sequential flow of information through the pipeline; question marks denote checklist item queries}
    \centering
    \label{fig:pipeline}
\end{figure}

We implemented a RAG pipeline to systematically apply the GRoLTS checklist to full-text research articles. The pipeline enables checklist-based appraisal by retrieving relevant textual evidence prior to answer generation. We use RAG to reduce reliance on the model’s memory by explicitly grounding responses in retrieved document content. The five stages of the pipeline as shown in Figure~\ref{fig:pipeline}: (1) Document acquisition and preprocessing: full-text articles are collected and converted into machine-readable format (Markdown). (2) Text segmentation: documents are divided into smaller textual chunks.(3) Embedding: each GRoLTS question and each chunk is transformed into a vector representation in embedding space.(4) Retrieval: for each GRoLTS question (represented by the question mark), the top-$k$ most semantically relevant chunks are retrieved based on vector similarity. (5) Prompt construction and answer generation: retrieved evidence is incorporated into a structured prompt, and the LLM generates a binary (\texttt{yes}/\texttt{no}) response. We will define each of these stages in detail below.

\paragraph{Document Acquisition and Preprocessing}
All full-text articles from the original GRoLTS study and additional case studies were downloaded as PDF files. Each PDF was converted into Markdown using the Python tool \texttt{Marker}~\citep{marker2025}. Markdown was used because it preserves structural elements such as headings and tables, thus facilitating more accurate retrieval and interpretation by LLMs.

\paragraph{Text Segmentation}
The Markdown documents were divided into overlapping chunks. Let $w_{\mathrm{chunk}}$ be the number of words per chunk and $w_{\mathrm{overlap}}$ the number of overlapping words between consecutive chunks. We set $w_{\mathrm{chunk}}\:=\:1000$ words and $w_{\mathrm{overlap}}\:=\:50$ words, yielding chunks of $w_{\mathrm{total}}\:=\:w_{\mathrm{chunk}}+w_{\mathrm{overlap}}\:=\:1050$ words. This chunk size was selected to remain within the input length limits of all embedding generation models considered, while the overlap preserved continuity across chunk boundaries and retained multi-sentence elements such as tables and extended methodological descriptions.

\paragraph{Embedding}
Let $t_{i,d}$ denote chunk $i$ from document $d$, and $q_n$ denote checklist item $n$. Both text chunks and checklist items were embedded in a shared vector space using \texttt{Qwen3-Embedding-8B}~\citep{qwen3embedding}. This model was selected because, at the time of experimentation, it demonstrated strong performance on the Massive Text Embedding Benchmark (MTEB), particularly for retrieval tasks involving long-form documents. Using a high-performing, general-purpose embedding model ensured robust semantic matching between checklist items and document segments without task-specific fine-tuning. The resulting vectors $\mathbf{e}_{i,d} \in \mathbb{R}^m$ (for text chunks) and $\mathbf{q}_n \in \mathbb{R}^m$ (for checklist items) encode semantic similarity in an $m$-dimensional space (e.g., $m=4096$), enabling retrieval of document segments most relevant to each checklist item.

\paragraph{Retrieval}
For each checklist item $q_n$ and document $d$, cosine similarity was computed between the question embedding $\mathbf{q}_n$ and each chunk embedding $\mathbf{e}_{i,d}$:
\begin{equation}
\mathrm{cos\_sim}(\mathbf{e}_{i,d}, \mathbf{q}_n) = \frac{\mathbf{e}_{i,d} \cdot \mathbf{q}_n} {\|\mathbf{e}_{i,d}\| \|\mathbf{q}_n\|}
\end{equation}
Chunks were ranked by similarity score, and the top $k = 10$ chunks were selected for each document–question pair. The highest-scoring chunks were concatenated and provided as contextual input to the LLM.

Selecting $k=10$ represented a balance between contextual coverage and computational efficiency, while remaining within the input length constraints of the LLMs. With a chunk size of 1,000 words and 50-word overlap, this corresponds to a maximum of approximately 10,500 words of retrieved context per checklist item.

\paragraph{Prompt Construction and Answer Generation}
A standardized prompt template (see Appendix~\ref{appendixC}) was used to integrate each GRoLTS question $q_n$ with its retrieved text chunks. The prompt specifies the extraction task, provides retrieved evidence as context, and constrains responses to a structured format consisting of: (1) a brief rationale, (2) quoted supporting evidence, and (3) a final binary judgment (YES/NO). The model was explicitly instructed to base its judgment solely on the provided context. The prompt was adapted from Bron et al.~\cite{bron2024} and refined through pilot testing to improve consistency of model outputs; after this stage, it was fixed and applied uniformly across all experiments. To minimize variability, all responses were generated using deterministic decoding (temperature = 0), with a fixed maximum output length of 1500 tokens. Repeated runs on a subset of document–question pairs confirmed stable outputs.

The pipeline was evaluated using multiple general-purpose LLMs: GPT-5-mini-2025-08-07~\citep{openai_gpt}, LLaMA 3.3 70B Instruct~\citep{meta_llama3_3_70b_instruct}, Qwen3 30B A3B Instruct 2507~\citep{qwen3technicalreport}, Qwen3-Next 80B A3B Instruct~\citep{qwen3technicalreport}, and Magistral Small 2509~\citep{mistralai2025magistral}. These models were selected because they are recent, high-capacity instruction-tuned LLMs that have demonstrated strong performance on information extraction and reasoning tasks relevant to research synthesis, while also being practically accessible through APIs or local deployment. GPT-5-mini was accessed via its API, and the other models were hosted locally, using identical prompt templates and retrieval settings to ensure comparability. Locally hosted models were run on Nvidia H100 GPUs with identical inference settings.

Prompts were processed in batches to improve computational efficiency while maintaining independent evaluation of each document–question pair by the LLM. Binary responses were extracted programmatically from the structured outputs, and all results were stored and indexed by document and checklist item for subsequent analysis. The complete set of prompts, model outputs, and annotations are publicly available at: \url{https://osf.io/kw8a7}.

\subsection{Evaluation and Analytic Strategy}
We conducted two primary comparisons aligned with the two-stage study design:

\begin{enumerate}
\item \textbf{Baseline human–LLM agreement (PTSD, v1):} 
We quantified the agreement between human annotations and LLM-generated responses for each question in the original GRoLTS checklist (v1) in the PTSD domain. This baseline establishes model performance prior to any checklist revision.

\item \textbf{Impact of checklist revision on human–LLM agreement (all domains, v2):} 
To evaluate whether the targeted revisions improved alignment, we compared human annotations with LLM responses using the revised checklist (v2) in all domains. This comparison assesses whether revising the checklist items reduces disagreement and improves consistency.
\end{enumerate}

The agreement between humans and LLMs was evaluated using three methods. (1) At the question level, the exact-match accuracy was computed between the human and each LLM annotation for each checklist item. (2) To account for the agreement expected by chance and the potential class imbalance, we computed Cohen's $\kappa$ for pairwise human–LLM agreement and Fleiss' $\kappa$ was used to assess the agreement between all raters (LLMs and the human annotator) for each checklist item~\citep{cohen1960,Fleiss1971}. (3) The total checklist scores were calculated per study and the resulting human and LLM ranking of the studies were analyzed using Spearman's rank correlation~\citep{spearman1904}.

\section{Results}

\subsection{Human-LLM Agreement: GRoLTS v1}
\begin{figure}
    \includegraphics[width=\linewidth, alt={Scatter plot of mean human-LLM agreement per GRoLTS v1 checklist question, ranked left to right from lowest to highest agreement (question 18 lowest, question 17 highest). Blue points show mean agreement for the PTSD case study with vertical error bars indicating standard deviation; gray points show individual model-level agreement scores. Agreement rises roughly monotonically from about 0.1 at the lowest-ranked question to near 1.0 at the highest-ranked questions, with PTSD means generally tracking the overall ranking but with notable spread among individual models at low- and mid-ranked questions.}]{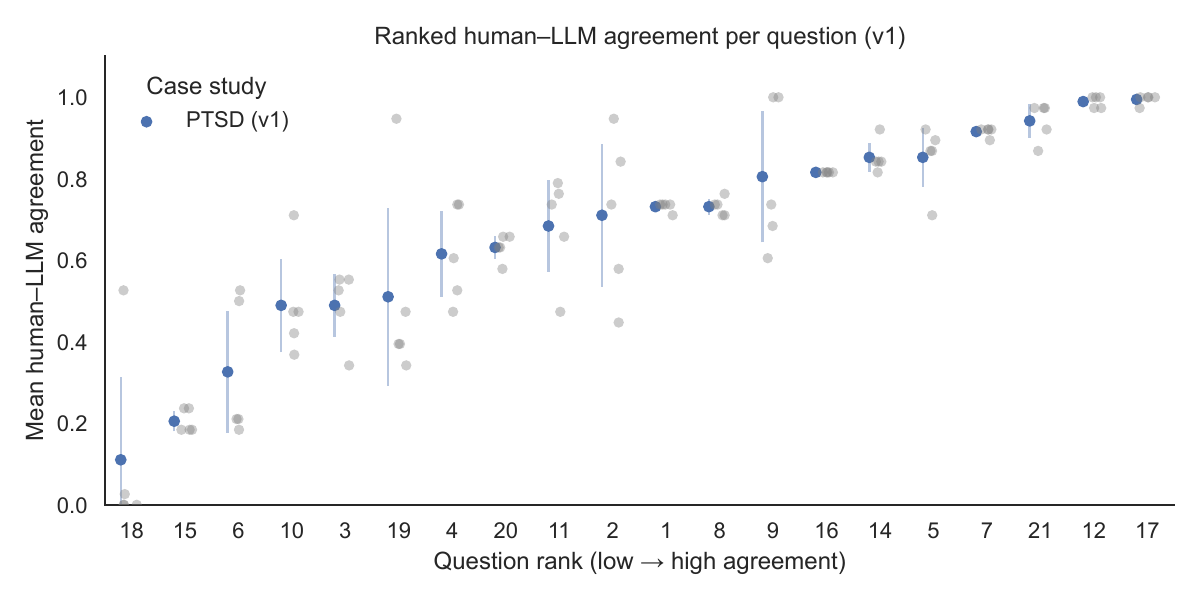}
    \caption{Mean human–LLM agreement for each question in GRoLTS v1 ranked from lowest to highest agreement along the x-axis. Blue points indicate the mean agreement for the PTSD case study, with vertical error bars showing the standard deviation. Gray points represent individual model-level agreement scores for each question}
    \centering
    \label{fig:ranked_v1}
\end{figure}

Figure~\ref{fig:ranked_v1} presents human–LLM agreement using exact-match accuracy between the human and each LLM for each question in GRoLTS v1, ordered from lowest to highest mean accuracy. The blue points indicate the mean accuracy across LLMs, error bars show the standard deviation across LLMs, and the gray points represent individual LLM accuracies. A complete overview of the accuracy per question and LLM is provided in Appendix~\ref{appendixD}.

Two main patterns emerge. First, the mean accuracy varies widely across questions, ranging from approximately 0.10 (question 18) to values approaching 1.0. Second, questions with lower mean accuracy tend to show greater variability between LLMs, as reflected in larger standard deviations and more dispersed individual scores, whereas higher-accuracy questions show tighter clustering for all models. These patterns suggest that some items are interpreted inconsistently across LLMs, whereas others yield more stable responses. Overall, performance is strongly question-dependent rather than uniform across the checklist, with substantial heterogeneity at the item level. Several questions show consistently low to moderate accuracy for all LLMs; these questions were used to guide targeted revisions, described in the next section.

\subsection{Item-level Challenges and Checklist Revisions}
Investigation of questions with low accuracy or high standard deviation from GRoLTS v1 identified recurring sources of disagreement that informed the checklist revisions. Two main issues were observed: (1) insufficient differentiation between partial and complete reporting, and (2) broad phrasing that allowed multiple interpretations.

To address the first issue, several items were split to distinguish between partial and complete reporting. For example, item 15 in v1 (``\textit{Are the number of cases per class reported for all models tested (absolute sample size or proportion)?}'') required identifying the total number of models and verifying whether class sizes were reported for each. Inspection of the LLM-generated explanations indicated that LLMs often identified class sizes for one or more models, but did not consistently verify whether this information was reported for all tested models. Consequently, LLMs often identified class sizes for a subset of models and responded \texttt{True}, whereas human raters marked the item as \texttt{False} when reporting was incomplete. In GRoLTS v2 (Appendix~\ref{appendixB}), this item was divided into two questions to separate these cases: one focusing on reporting for the final model and one on reporting across all tested models. A similar approach was applied to item 14 (v1), which was split into items 11 and 12 (v2).

A second set of revisions addressed ambiguous phrasing. For example, item 18 (v1) asked whether plots of estimated mean trajectories were included for each model. LLMs frequently identified the presence of at least one plot and responded \texttt{True}, whereas human raters required plots for each model. This item was omitted in v2 due to persistent ambiguity and low reporting prevalence (Appendix~\ref{appendixD}). Additional items (3, 4, and 9) were removed following similar concerns.

Finally, several items were revised to make them more specific. For instance, item 6 (v1) (``\textit{Is information about the distribution of the observed variables included?}'') was often marked \texttt{True} by LLMs when minimal descriptive statistics were reported, whereas human raters required evidence of distributional assumptions or testing. In v2, this item was reformulated with explicit examples (e.g., tests for within-class normality or multivariate normality). Similar clarifications were applied to items 1, 5, 6, and 10 (v1), resulting in items 1, 3, 4, and 7 in v2. 

Overall, the revisions aimed to reduce ambiguity, distinguish partial from complete reporting, and make the evaluation criteria more explicit, thus improving the alignment between human and LLM assessments.

\subsection{Human-LLM Agreement: GRoLTS v2}
\begin{figure}
    \includegraphics[width=\linewidth, alt={Scatter plot of mean human-LLM agreement per GRoLTS v2 checklist question, ranked left to right from lowest to highest agreement (question 14 lowest, question 16 highest). Blue, orange, and green points show mean agreement with vertical error bars for the Achievement, Delinquency, and PTSD v2 case studies respectively; gray points show individual model-level agreement scores. Agreement rises roughly monotonically from about 0.3-0.4 at the lowest-ranked question to near 1.0 at the highest-ranked questions, with the three case studies tracking closely together across most questions and overall higher and less variable agreement than seen in the v1 checklist.}]{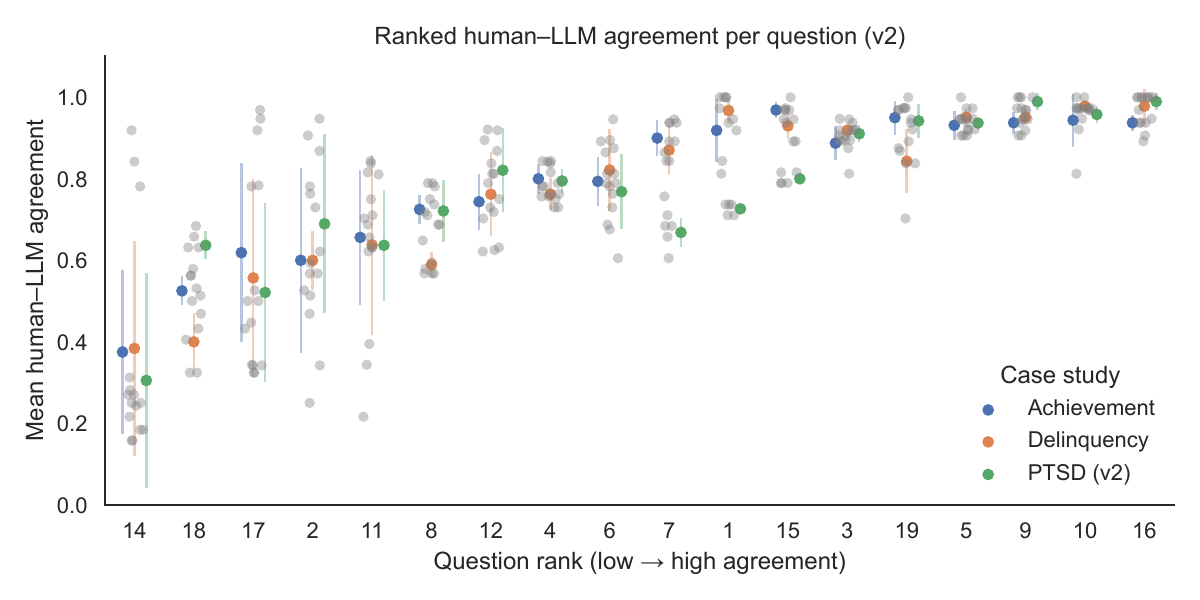}
    \caption{Mean human–LLM agreement for each question in GRoLTS v2, ranked from lowest to highest agreement along the x-axis. Colored points denote mean agreement for each case study (Achievement, Delinquency, PTSD), with vertical error bars showing the standard deviation. Gray points represent individual model-level agreement scores for each question}
    \centering
    \label{fig:ranked_v2}
\end{figure}

Figure~\ref{fig:ranked_v2} presents human–LLM agreement using exact-match accuracy between the human and each LLM for each question in GRoLTS v2 (Appendix~\ref{appendixB}), sorted by mean accuracy. A full overview of these results is provided in Appendix~\ref{appendixE}. Several differences emerge compared to GRoLTS v1.

Overall, agreement is higher in v2. In v1, the lowest-ranked questions showed mean accuracies of approximately 0.10 to 0.20, whereas in v2 the lowest values are generally in the range of 0.40 to 0.60, indicating a reduction in very low-performing questions. A larger proportion of items now fall in the higher accuracy range (0.7–1.0), with several questions approaching perfect agreement. The variability across LLMs is also reduced for many mid- and high-performing items: whereas v1 showed substantial error bars for several mid-range questions, v2 exhibits tighter clustering for higher-accuracy items, although some variability remains.

The inclusion of multiple case studies (Achievement, Delinquency, PTSD (v2)) further enables evaluation of cross-domain consistency. Agreement patterns are similar across case studies for higher-ranked items, suggesting that improvements generalize beyond a single domain. Greater divergence is observed among lower-ranked items, indicating that some questions remain sensitive to interpretation. In particular, PTSD (v2) shows deviations for questions 7, 1, and 15; this may reflect artifacts introduced when mapping the v1 human annotations to the revised v2 structure, or differences in how the revised items were interpreted by human raters. Overall, these results indicate an upward shift in accuracy and reduced variability following the checklist revisions, particularly among previously low- and mid-performing items, although some question-level challenges remain.

\subsection{Chance-Corrected Agreement}
The previous results demonstrate a clear upward shift in human–LLM agreement from v1 to v2. We next assess whether these improvements persist after correcting for the agreement expected by chance. To quantify chance-corrected agreement, we computed Cohen’s $\kappa$ between each LLM and human annotations across the three GRoLTS v2 datasets (Achievement, Delinquency, PTSD (v2)), as well as for the original GRoLTS v1 PTSD dataset (Table~\ref{tab:cohens-kappa}). Following commonly used interpretive guidelines (e.g., Landis and Koch~\citep{Landis1977}), $\kappa$ values between 0.41 and 0.60 are typically described as moderate agreement and values between 0.61 and 0.80 as substantial agreement.

For GRoLTS v2, Cohen’s $\kappa$ ranged from 0.41 to 0.72 across the datasets and LLMs, indicating moderate to substantial agreement. GPT-5 mini achieved the highest agreement with human annotations, while the differences among the remaining models were relatively small. The $\kappa$ values for the original PTSD v1 dataset were lower (0.31–0.52), consistent with the weaker item-level agreement observed earlier.

\begin{table}[t]
    \caption{Cohen's $\kappa$ between human and LLM annotations}
    \centering
    \begin{tabular}{lllll}
    \toprule
        Model & Achievement & Delinquency & PTSD (v2) & PTSD (v1) \\
        \midrule
        Llama 3.3 70B & 0.61 & 0.54 & 0.52 & 0.40 \\
        Magistral Small & 0.51 & 0.41 & 0.42 & 0.33 \\
        Qwen 3 30B & 0.55 & 0.52 & 0.46 & 0.31\\
        Qwen Next 80B & 0.54 & 0.51 & 0.48 & 0.33\\
        GPT-5 mini & 0.67 & 0.72 & 0.73 & 0.52 \\
    \bottomrule
    \end{tabular}
    \label{tab:cohens-kappa}
\end{table}

We computed Fleiss’ $\kappa$ for all raters (five LLMs and one human expert) for each dataset (Table~\ref{tab:fleiss-kappa}) to evaluate multi-rater consistency. Fleiss’ $\kappa$ was consistently around 0.60 to 0.61 for GRoLTS v2, compared to 0.41 for the original PTSD v1 checklist. These results indicate a higher overall consistency among LLMs and the human rater in v2 relative to v1.

\begin{table}[t]
    \caption{Fleiss' $\kappa$ across all raters (five LLMs and one human expert)}
    \centering
    \begin{tabular}{ll}
    \toprule
        Dataset & Fleiss' $\kappa$ \\
        \midrule
        Achievement & 0.60 \\
        Delinquency & 0.61 \\
        PTSD (v2) & 0.60 \\
        PTSD (v1) & 0.41 \\
    \bottomrule
    \end{tabular}
    \label{tab:fleiss-kappa}
\end{table}

\subsection{Rank-order Consistency}
Beyond item-level agreement, we evaluated whether LLM-generated scores preserve the relative order of studies produced by human appraisal. This analysis addresses the practical utility of LLM-assisted checklist scoring: even when some items are incorrectly scored, models may still correctly prioritize studies by overall reporting quality.

The overall rank consistency was assessed using Spearman’s rank correlation coefficient $\rho$, which evaluates whether two raters preserve relative ordering without considering absolute score differences~\citep{spearman1904}. This metric is particularly relevant for checklist-based appraisal, where decisions often depend on prioritization rather than exact score equivalence. Table~\ref{tab:spearman-rank} shows that correlations ranged from 0.36 to 0.72 across datasets and LLMs, indicating moderate to strong rank-order agreement depending on the model and dataset.

\begin{table}[t]
    \caption{Overall rank correlations (Spearman $\rho$) between human and LLM total scores}
    \centering
    \begin{tabular}{lllll}
    \toprule
        Model & Achievement & Delinquency & PTSD (v2) & PTSD (v1) \\
        \midrule
        Llama 3.3 70B & 0.72 & 0.56 & 0.56 & 0.47 \\
        Magistral Small & 0.46 & 0.39 & 0.57 & 0.46 \\
        Qwen 3 30B & 0.46 & 0.53 & 0.57 & 0.43\\
        Qwen Next 80B & 0.36 & 0.45 & 0.65 & 0.59\\
        GPT-5 mini & 0.56 & 0.65 & 0.41 & 0.63 \\
    \bottomrule
    \end{tabular}
    \label{tab:spearman-rank}
\end{table}

\begin{figure}
    \includegraphics[width=0.8\linewidth, alt={Grid of four bar charts, one per case study (Achievement, Delinquency, PTSD v2, PTSD v1), showing Spearman rank correlation between human and LLM scores for five models (gpt-5-mini, Qwen-3-30B, Llama-3.3-70B, Magistral-Small, Qwen-Next-80B). Within each panel, bars are grouped by model and colored by checklist item agreement level (high, mid, low). Correlations are generally highest for high-agreement items and lowest, sometimes near zero or slightly negative, for low-agreement items, with this pattern holding across most models and case studies.}]{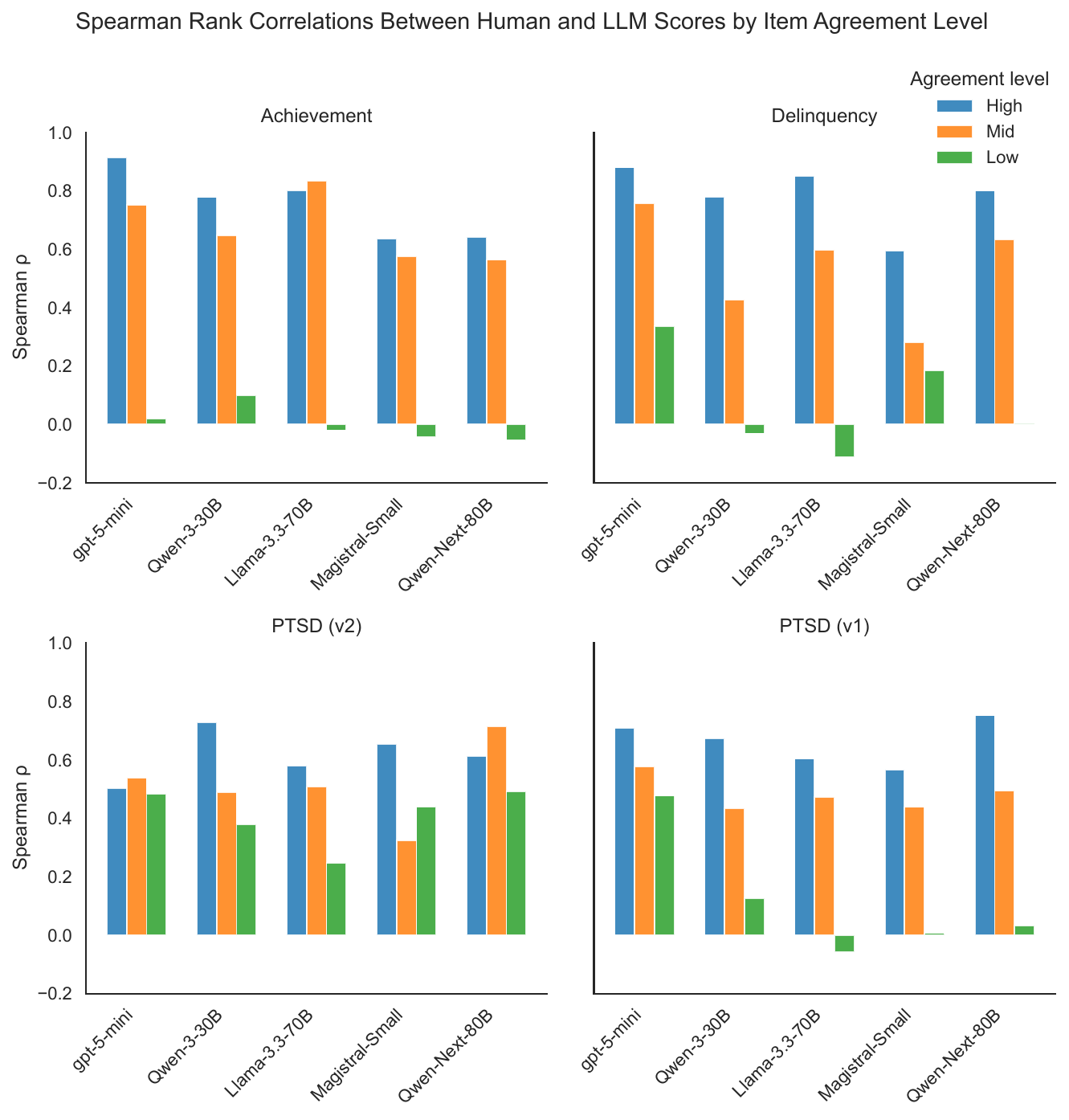}
    \caption{Spearman rank correlations ($\rho$) between human total checklist scores and LLM-generated total scores, computed separately using subsets of checklist items categorized by prior human–LLM agreement (high, mid, and low). Results are shown for four datasets: Achievement, Delinquency, PTSD (v2), and the original PTSD (v1)}
    \centering
    \label{fig:rank-order-difficulty}
\end{figure}

We further examined the sensitivity of rankings to question agreement by recomputing total scores using subsets of items grouped by prior human–LLM agreement. Figure~\ref{fig:rank-order-difficulty} shows that rankings derived from high-agreement items aligned well with human-derived rankings ($\rho$ $\approx$ 0.58–0.93), whereas rankings based on middle-agreement items showed moderate consistency ($\rho$ $\approx$ 0.47–0.76). Rankings derived from low-agreement items were substantially less stable, in some cases approaching zero or becoming negative.

These results indicate that ranking stability is strongly influenced by item-level agreement: retaining high-agreement items leads to substantial correlation with human rankings, whereas including low-agreement items reduces stability. These findings suggest the viability of a selective-automation strategy in which LLMs are used for well-specified criteria, while human reviewers focus on ambiguous or conceptually complex items.

\section{Discussion}

This study evaluated the extent to which LLMs can approximate expert human judgments in checklist-based quality appraisal. Across multiple case studies and checklist versions, we observed moderate to substantial agreement between LLM-generated and human annotations. Targeted revisions to the checklist, informed by patterns of disagreement in GRoLTS v1, led to improvements in both raw and chance-corrected agreement. These findings indicate that the reliability of LLM-assisted appraisal depends not only on model capability, but also on the clarity and structure of the appraisal instrument. Furthermore, they suggest that human–LLM disagreement can serve as a useful diagnostic signal for identifying ambiguous or underspecified checklist items. Our findings complement recent studies showing that LLMs can support several stages of the evidence synthesis workflow, including screening, data extraction, and risk-of-bias assessment~\citep{Dennstadt2024, Khraisha2024, Li2026}. Taken together, this work shows that disagreement between human and automated assessments can be used not only to evaluate performance, but also to improve the design of appraisal instruments.

\subsection{Question-Level Agreement as a Diagnostic Tool}
A key insight of this work is that LLM performance is highly question-dependent. Some checklist questions achieved near-perfect agreement, whereas others showed consistent disagreement and variability. The most problematic items were not random but shared recurring characteristics, including conditional phrasing (e.g., ``for all models tested''), ambiguous scope definitions, and broadly defined conceptual criteria without explicit guidance. Similar sources of variability have also been observed among human reviewers applying appraisal tools, where ambiguous or conditional criteria often reduce inter-rater reliability~\citep{Hartling2013, ArmijoOlivo2012}.

The LLMs often extracted relevant information from the retrieved text but did not consistently distinguish whether reporting was complete or whether the criterion was fully satisfied. These patterns provided the basis for targeted checklist revisions. The improvements observed for the GRoLTS v2 checklist suggest that systematic analysis of human–LLM disagreement can help identify ambiguous or underspecified checklist items and guide the revision of reporting instruments.

\subsection{Implications for LLM-Assisted Quality Appraisal}
Beyond item-level improvements, our findings demonstrate that LLM-assisted checklist appraisal can achieve moderate to substantial alignment with human experts when using RAG. Importantly, the improvements introduced in GRoLTS v2 were not limited to raw accuracy but extended to chance-corrected agreement, suggesting that a clearer checklist design may benefit both human and automated annotators.

The rank-order analyses further clarify the practical utility of LLM-based appraisal. Even when some individual items are misclassified, LLMs can preserve the relative ordering of studies, particularly when high-agreement items are retained. This indicates that LLM-assisted scoring may be sufficient for prioritization and screening tasks, provided that ambiguous criteria are reviewed by human experts.

At the same time, differences between LLMs were present, but not dominant. Although some models (e.g., GPT-5 mini) consistently achieved higher agreement with human annotations, the differences among the remaining models were comparatively modest. This suggests that, beyond model choice, performance is also influenced by factors such as the retrieved information and checklist design. These findings highlight that effective automation in research synthesis depends not only on model capability, but also on retrieval strategies and the design of appraisal instruments.

\subsection{Limitations and Future Work}
Several limitations should be acknowledged. First, LLM outputs remain sensitive to prompt formulation and instruction framing, which may introduce variability in item-level responses. These results are consistent with earlier work showing that LLM performance in evidence synthesis tasks often depends strongly on question formulation and prompt structure~\citep{Chang2024, anagnostidis2024}. Second, human annotations themselves are subject to inter-rater variability and conceptual ambiguity, which may place an upper bound on achievable agreement between models and expert raters.

Third, the checklist revisions and evaluations were conducted across a limited number of datasets. Although improvements generalized across the included case studies, additional datasets may reveal domain-specific reporting practices or ambiguities not captured here. Finally, parameters of the RAG pipeline were selected pragmatically rather than exhaustively optimized; alternative chunking strategies, retrieval thresholds, or prompt structures may yield additional improvements.

Future work should therefore explore iterative refinement of appraisal instruments through structured human–LLM collaboration. Expanding evaluations across additional domains and larger corpora will help determine the robustness of these findings. Integrating LLM-assisted appraisal into real-world review workflows where models generate preliminary assessments and human reviewers resolve ambiguous cases represents a promising direction.

\section*{Declarations}
\paragraph{Funding}
This work was supported by the Dutch Research Council (NWO) using grant number VI.C.231.102. This work used the Dutch national e-infrastructure with the support of the SURF Cooperative using grant no. EINF-11739.

\paragraph{Competing Interests}
None

\paragraph{Ethics approval}
Not applicable

\paragraph{Consent to participate}
Not applicable

\paragraph{Consent for publication}
Not applicable

\paragraph{Availability of data and materials}
Data can be found on OSF: \url{https://osf.io/kw8a7}. 

\paragraph{Code availability}
Code can be found on GitHub: \url{https://github.com/timovdk/grolts-llm}.

\paragraph{Author Contributions}
Conceptualization: T.K.; M.Z.; R.S.; B.G.; E.G. Methodology: T.K.; R.B.; R.S.; E.G. Formal analysis: T.K.; E.G. Investigation: T.K.; E.G. Software: T.K. Data curation: T.K.; B.C.; K.D.; B.G.; S.H.; S.R.; E.G. Resources: K.D.; B.G.; S.H.; S.R. Visualization: T.K.; R.B.; E.G. Project administration: E.G. Supervision: R.B.; R.S.; E.G. Funding acquisition: T.K.; R.S. Writing – original draft: T.K.; E.G. Writing – review \& editing: T.K.; B.C.; M.Z.; R.B.; R.S.; K.D.; B.G.; S.H.; S.R.; E.G. All authors approved the final submitted draft.

\bibliography{references}

\appendix
\section{Original GRoLTS Questions (v1)}\label{appendixA}
\begin{enumerate}
    \item Is the metric of time used in the statistical model reported?
    \item Is information presented about the mean and variance of time within a wave?
    \item Is the missing data mechanism reported?
    \item Is a description provided of what variables are related to attrition/missing data?
    \item Is a description provided of how missing data in the analyses were dealt with?
    \item Is information about the distribution of the observed variables included?
    \item Is the software mentioned?
    \item Are alternative specifications of within-class heterogeneity considered (e.g., LGCA vs. LGMM) and clearly documented?
    \item Are alternative specifications of the between-class differences in variancecovariance matrix structure considered and clearly documented?
    \item Are alternative shape/functional forms of the trajectories described?
    \item If covariates have been used, can analyses still be replicated?
    \item Is information reported about the number of random start values and final iterations included?
    \item Are the model comparison (and selection) tools described from a statistical perspective?
    \item Are the total number of fitted models reported, including a one-class solution?
    \item Are the number of cases per class reported for each model (absolute sample size, or proportion)?
    \item If classification of cases in a trajectory is the goal, is entropy reported?
    \item Is a plot included with the estimated mean trajectories of the final solution?
    \item Are plots included with the estimated mean trajectories for each model?
    \item Is a plot included of the combination of estimated means of the final model and the observed individual trajectories split out for each latent class?
    \item Are characteristics of the final class solution numerically described (i.e., means, SD/SE, n, CI, etc.)?
    \item Are the syntax files available (either in the appendix, supplementary materials, or from the authors)?
\end{enumerate}

\newpage
\section{Revised GRoLTS Questions (v2)}\label{appendixB}
\begin{enumerate}
    \item Is the metric or unit of time used in the statistical model reported (i.e., wave, hours, days, weeks, months, years, etc.)?
    \item Is information presented about the mean and variance of time within a wave?
    \item Is a description provided of how missing data in the analyses were dealt with (i.e., List wise deletion, multiple imputation, Full information maximum likelihood (FIML) etc.)?
    \item Is information about the distribution of the observed variables included (i.e., tests for normally distributed variables within classes, multivariate normality, etc.)?
    \item Is the software that was used for the statistical analysis mentioned?
    \item Are alternative specifications of within-class heterogeneity considered (e.g., LGCA vs. LGMM) and clearly documented?
    \item Are alternative shape/functional forms of the trajectories described (e.g., was it tested whether a quadratic trend or a non-linear form would fit the data better)?
    \item If covariates or predictors have been used, is it done in such a way that the analyses could be replicated?
    \item Is information reported about the number of random start values and final iterations included?
    \item Are the model comparison (and selection) tools described from a statistical perspective?
    \item Are the total number of fitted models reported?
    \item Is information about a one-class solution reported?
    \item Are the number of cases per class reported for the final model (absolute sample size, or proportion)?
    \item Are the number of cases per class reported for the all models tested (absolute sample size, or proportion)?
    \item Is entropy reported?
    \item Is a plot included with the estimated mean trajectories of the final solution?
    \item Is a plot included of the combination of estimated means of the final model and the observed individual trajectories split out for each latent class?
    \item Are characteristics of the final class solution numerically described (i.e., means, SD/SE, n, CI, etc.)?
    \item Are the syntax files available (either in the appendix, supplementary materials, or from the authors)?
\end{enumerate}

\newpage
\section{Prompt Template}\label{appendixC}
{\small
\begin{verbatim}
SYSTEM_PROMPT = """
    You are evaluating whether an academic paper reports specific 
    methodological or statistical information.

    Answer the QUESTION using only the given CONTEXT, which is 
    markdown-formatted text from a single academic paper.

    Output exactly in the following format:

    REASONING:
    Explain briefly how the answer follows from the CONTEXT.
    Use only information explicitly stated or logically implied. 
    Do not speculate or rely on typical practices in the field.

    EVIDENCE:
    - Quote exact phrases or sentences from the CONTEXT that
    support the reasoning, one per line.
    - Quotes must be verbatim excerpts from the CONTEXT.
    - If no direct quotes support the conclusion, leave this
    section empty.

    ANSWER:
    YES or NO

    Rules:
    - Use only the provided CONTEXT. Do not use outside knowledge.
    - Treat references to supplements, appendices, datasets, or
    URLs mentioned in the CONTEXT as valid and available.
    - Answer YES only when explicit evidence is present in the
    CONTEXT.
    - If the information is missing, unclear, or only implied by
    general practice, answer NO.
    - Do not add any text before or after the specified format.
    """

USER_PROMPT = """
    QUESTION: {question}
    CONTEXT: {context}
    """
\end{verbatim}
}

\newpage
\section{Item-level accuracy per LLM for GRoLTS v1 on PTSD}\label{appendixD}
\begin{figure}[ht]
    \includegraphics[width=\linewidth, alt={Heatmap of per-question accuracy for PTSD v1, with 21 checklist questions as rows and five models (gpt-5-mini, Qwen-3-30B, Llama-3.3-70B, Magistral-Small, Qwen-Next-80B) as columns, color-coded from dark (low accuracy) to light (high accuracy) with numeric values overlaid. A row-mean column on the right shows per-question accuracy averaged across models, and a bottom row shows per-model accuracy averaged across questions, with an overall mean of 0.68. A separate gray-scale column on the far right shows the proportion of human-coded 1's per question, for reference against accuracy. Accuracy is generally high (above 0.7) for most questions and models, but several questions, notably 6, 15, and especially 18, show consistently low accuracy across all models.}]{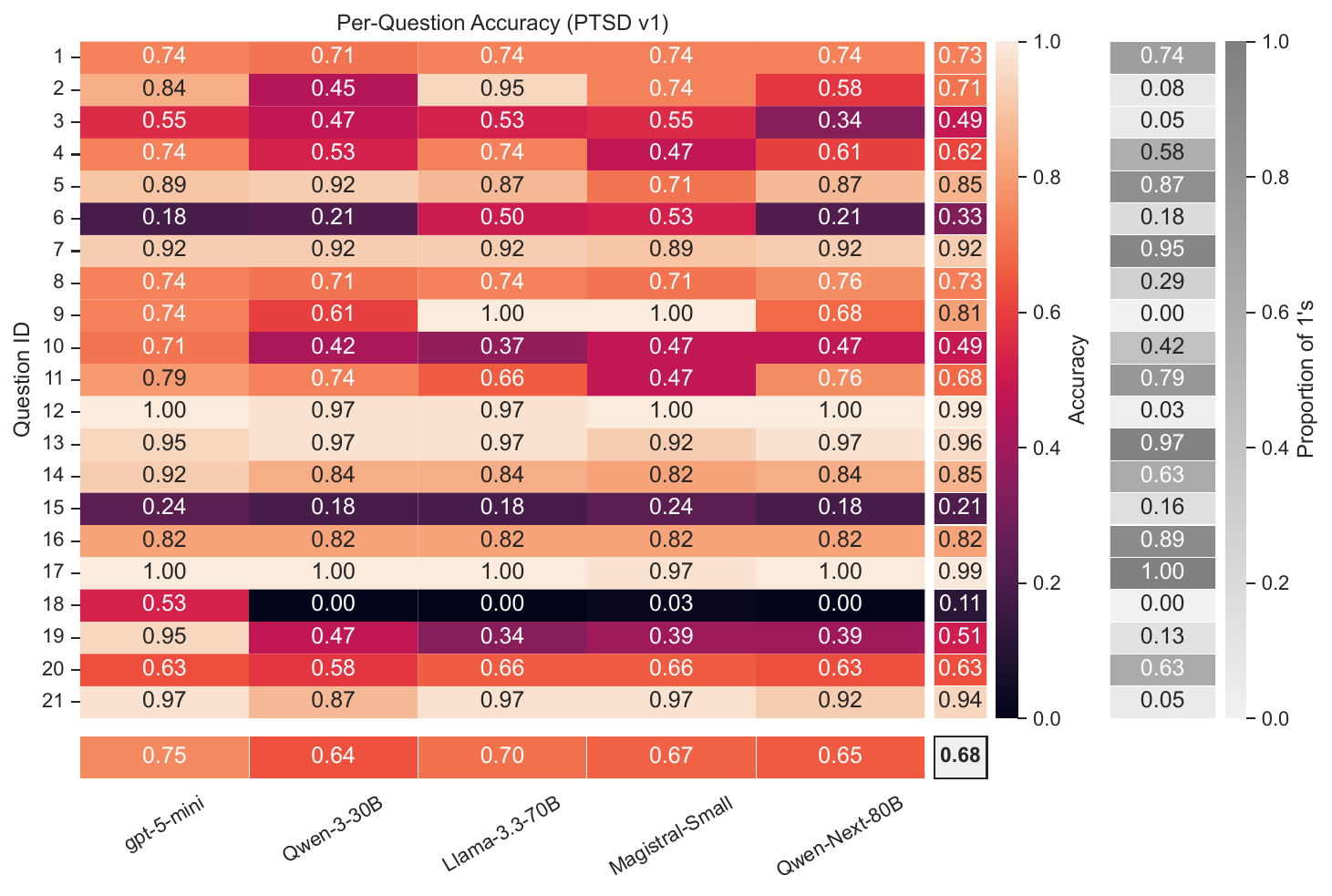}
    \caption{Item-level accuracy of five LLMs on the GRoLTS v1 checklist for the PTSD use case. Rows represent checklist items (Q1–Q21) and columns represent models; cell values indicate agreement with the human labels. The bottom row shows mean accuracy per model, the last column shows the mean accuracy per question, and the right panel displays the proportion of positive (``1'') labels per item}
    \centering
    \label{fig:item-level-accuracy}
\end{figure}

\newpage
\section{Item-level accuracy per LLM for GRoLTS v2 on Achievement, Delinquency, and PTSD}\label{appendixE}
\begin{figure}[ht]
    \includegraphics[width=\linewidth, alt={Heatmap of per-question accuracy for the Achievement case study, with 19 checklist questions as rows and five models (gpt-5-mini, Qwen-3-30B, Llama-3.3-70B, Magistral-Small, Qwen-Next-80B) as columns, color-coded from dark (low accuracy) to light (high accuracy) with numeric values overlaid. A row-mean column on the right shows per-question accuracy averaged across models, and a bottom row shows per-model accuracy averaged across questions, with an overall mean of 0.79. A separate gray-scale column on the far right shows the proportion of human-coded 1's per question, for reference against accuracy. Accuracy is generally high (above 0.7) for most questions and models, but questions 2, 14, 17, and 18 show notably lower and more variable accuracy across models.}]{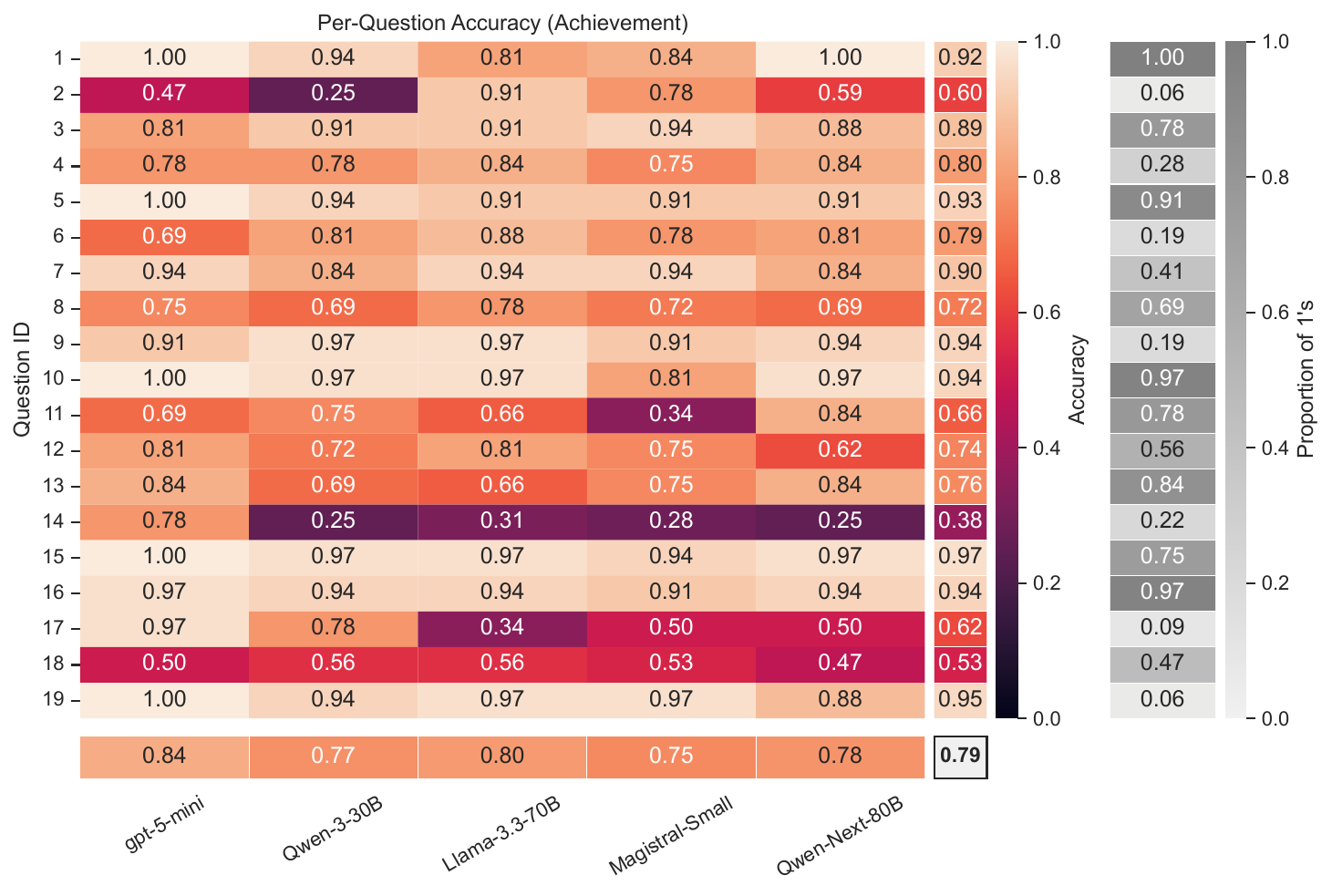} 
    \caption{Item-level accuracy of five LLMs on the GRoLTS v2 checklist for the Educational Achievement use case. Rows represent checklist items (Q1–Q19) and columns represent models; cell values indicate agreement with the human labels. The bottom row shows mean accuracy per model, the last column shows the mean accuracy per question, and the right panel displays the proportion of positive (``1'') labels per item}
    \centering
    \label{fig:item-level-accuracy_v2_ach}
\end{figure}
\begin{figure}[ht]
    \includegraphics[width=\linewidth, alt={Heatmap of per-question accuracy for the Delinquency case study, with 19 checklist questions as rows and five models (gpt-5-mini, Qwen-3-30B, Llama-3.3-70B, Magistral-Small, Qwen-Next-80B) as columns, color-coded from dark (low accuracy) to light (high accuracy) with numeric values overlaid. A row-mean column on the right shows per-question accuracy averaged across models, and a bottom row shows per-model accuracy averaged across questions, with an overall mean of 0.78. A separate gray-scale column on the far right shows the proportion of human-coded 1's per question, for reference against accuracy. Accuracy is generally high (above 0.7) for most questions and models, but questions 2, 8, 14, 17, and 18 show notably lower and more variable accuracy across models.}]{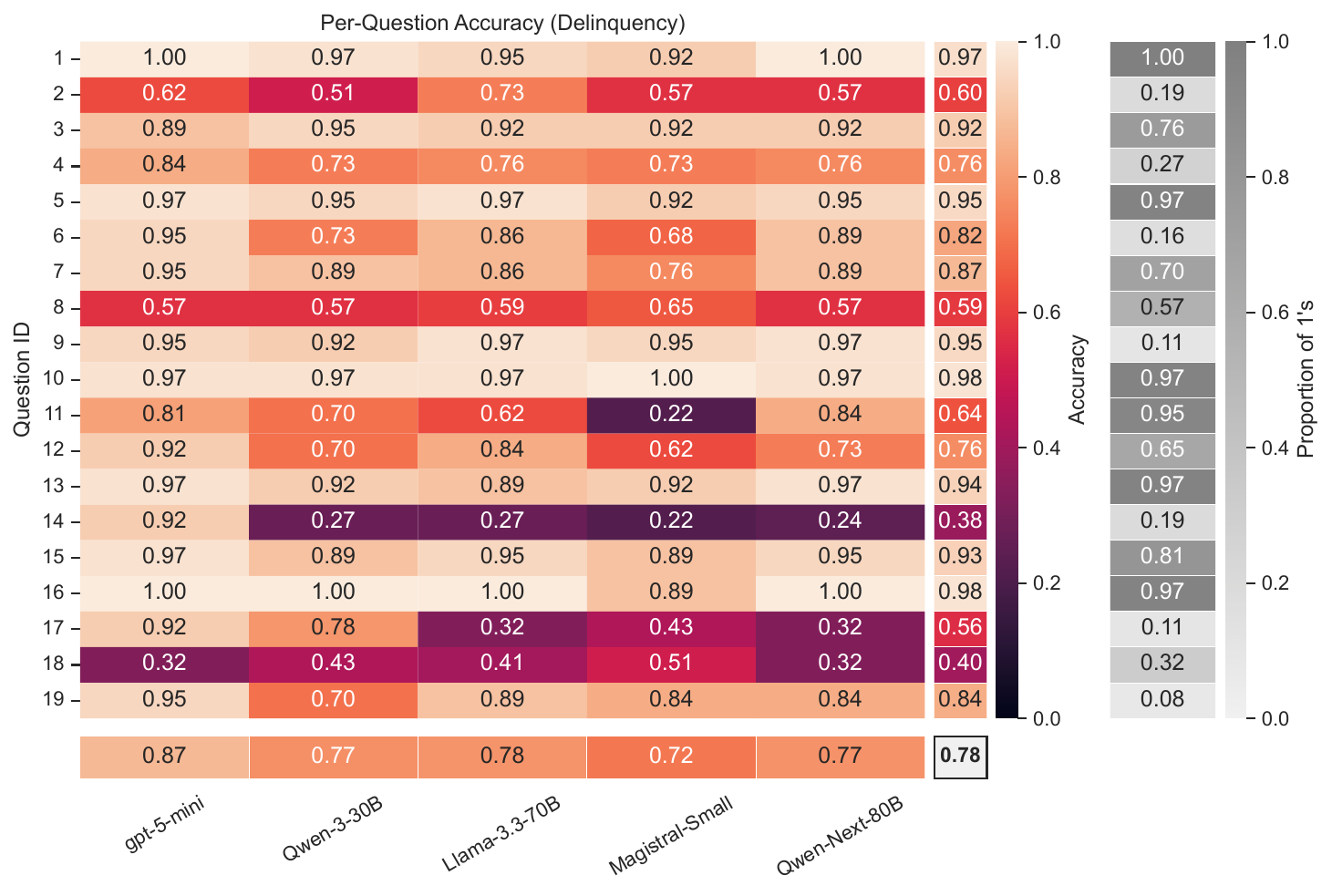}   
    \caption{Item-level accuracy of five LLMs on the GRoLTS v2 checklist for the Adolescent Delinquency use case. Rows represent checklist items (Q1–Q19) and columns represent models; cell values indicate agreement with the human labels. The bottom row shows mean accuracy per model, the last column shows the mean accuracy per question, and the right panel displays the proportion of positive (``1'') labels per item}
    \centering
    \label{fig:item-level-accuracy_v2_del}
\end{figure}
\begin{figure}[ht]
    \includegraphics[width=\linewidth, alt={Heatmap of per-question accuracy for PTSD v2, with 19 checklist questions as rows and five models (gpt-5-mini, Qwen-3-30B, Llama-3.3-70B, Magistral-Small, Qwen-Next-80B) as columns, color-coded from dark (low accuracy) to light (high accuracy) with numeric values overlaid. A row-mean column on the right shows per-question accuracy averaged across models, and a bottom row shows per-model accuracy averaged across questions, with an overall mean of 0.74. A separate gray-scale column on the far right shows the proportion of human-coded 1's per question, for reference against accuracy. Accuracy is generally high (above 0.7) for most questions and models, but questions 7, 11, 13, 14, and 17 show notably lower and more variable accuracy across models, with question 13 the lowest overall (mean 0.16).}]{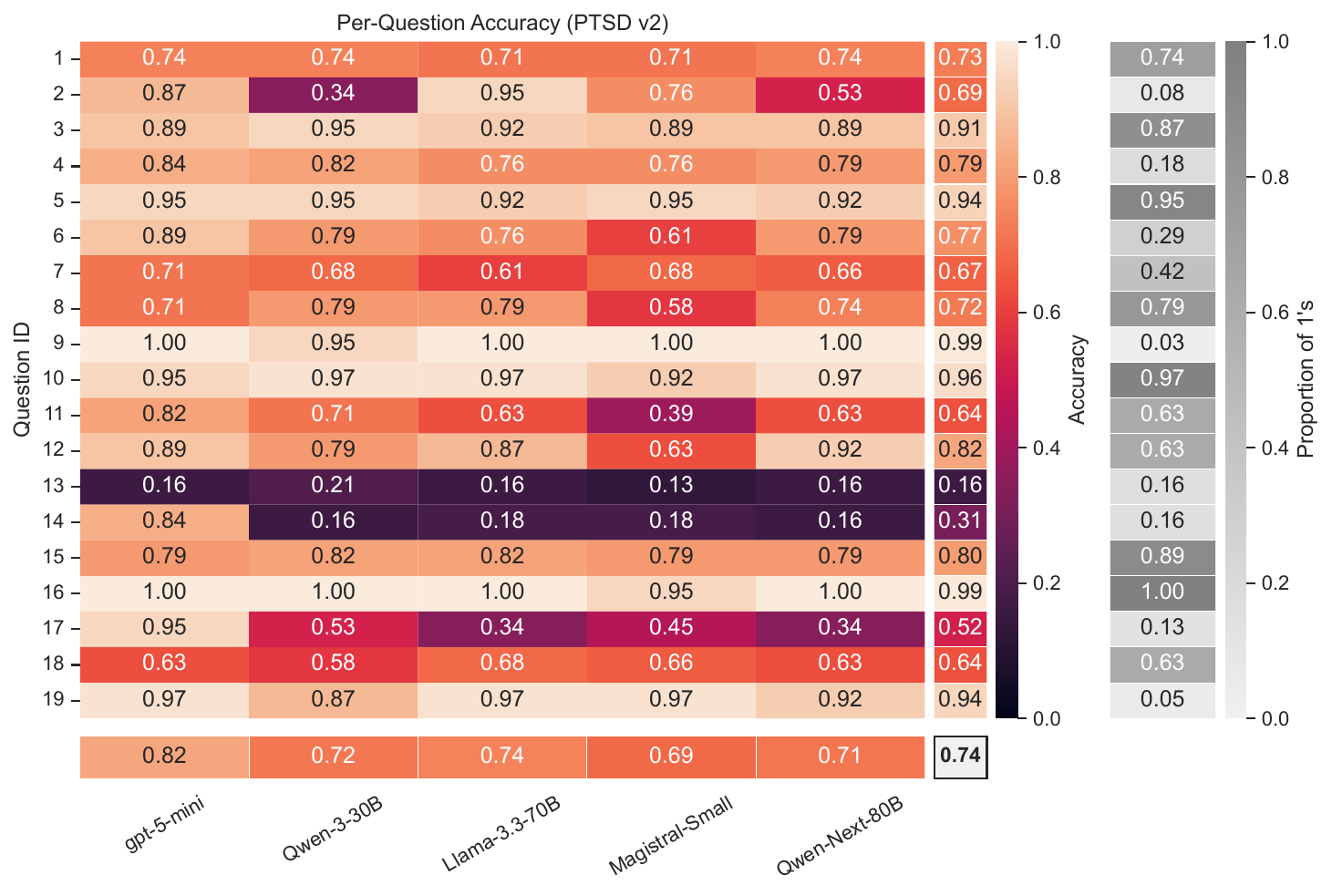}
    \caption{Item-level accuracy of five LLMs on the GRoLTS v2 checklist for the PTSD use case. Rows represent checklist items (Q1–Q19) and columns represent models; cell values indicate agreement with the human labels. The bottom row shows mean accuracy per model, the last column shows the mean accuracy per question, and the right panel displays the proportion of positive (``1'') labels per item}
    \centering
    \label{fig:item-level-accuracy_v2_ptsd}
\end{figure}

\end{document}